\documentclass{article} 
\usepackage{iclr_fmwild,times}

\usepackage{amsmath,amsfonts,bm}

\def\eqref#1{equation~\ref{#1}}

\def\1{\bm{1}}

\DeclareMathAlphabet{\mathsfit}{\encodingdefault}{\sfdefault}{m}{sl}
\SetMathAlphabet{\mathsfit}{bold}{\encodingdefault}{\sfdefault}{bx}{n}

\usepackage{listings}

\usepackage{xcolor}
\usepackage{authblk}

\definecolor{codegreen}{rgb}{0,0.6,0}
\definecolor{codegray}{rgb}{0.5,0.5,0.5}
\definecolor{codepurple}{rgb}{0.58,0,0.82}
\definecolor{backcolour}{rgb}{0.95,0.95,0.92}

\lstdefinestyle{mystyle}{
  backgroundcolor=\color{backcolour}, commentstyle=\color{codegreen},
  keywordstyle=\color{magenta},
  numberstyle=\tiny\color{codegray},
  stringstyle=\color{codepurple},
  basicstyle=\ttfamily\tiny,
  breakatwhitespace=false,         
  breaklines=true,                 
  captionpos=b,                    
  keepspaces=true,                 
  numbers=none,                    
  numbersep=5pt,                  
  showspaces=false,                
  showstringspaces=false,
  showtabs=false,                  
  tabsize=2
}

\usepackage{hyperref}
\usepackage{placeins}  
\usepackage{url}
\usepackage{booktabs}

\usepackage{graphicx} 
\usepackage{caption} 
\usepackage{listings}
\usepackage{algorithm}
\usepackage{algpseudocode}
\usepackage{graphicx}
\usepackage{subcaption} 
\title{Exploring LLM Agents for Cleaning Tabular Machine Learning Datasets}

\author{
    \textbf{Tommaso Bendinelli}$^{1,2}$ \quad
    \textbf{Artur Dox}$^{2}$ \quad
    \textbf{Christian Holz}$^{1}$ \quad\\
    $^1$Department of Computer Science, ETH Zurich, Switzerland \\
    $^2$ CSEM SA, Alpnach, Switzerland\\
}

\iclrfinalcopy 
\begin{document}

\maketitle

\begin{abstract}

High-quality, error-free datasets are a key ingredient in building reliable, accurate, and unbiased machine learning (ML) models. However, real world datasets often suffer from errors due to sensor malfunctions, data entry mistakes, or improper data integration across multiple sources that can severely degrade model performance.
Detecting and correcting these issues typically require tailor-made solutions and demand extensive domain expertise. Consequently, automation is challenging, rendering the process labor-intensive and tedious. In this study, we investigate whether Large Language Models (LLMs) can help alleviate the burden of manual data cleaning. We set up an experiment in which an LLM, paired with Python, is tasked with cleaning the training dataset to improve the performance of a learning algorithm without having the ability to modify the training pipeline or perform any feature engineering.
We run this experiment on multiple Kaggle datasets that have been intentionally corrupted with errors. Our results show that LLMs can identify and correct erroneous entries—such as illogical values or outliers—by leveraging contextual information from other features within the same row, as well as feedback from previous iterations. However, they struggle to detect more complex errors that require understanding data distribution across multiple rows, such as trends and biases.

\end{abstract}

\section{Introduction}
Despite being essential for achieving high model performance \citep{jager2024data}, data cleaning remains one of the least engaging yet most time-consuming tasks for data scientists  \citep{sambasivan2021everyone}. While many algorithms and approaches exist to speed up this process, significant challenges remain in efficiently deploying fully automated methods in real world scenarios \citep{ni2023automatic}. This is because devising a one-size-fits-all method that works universally for data cleaning is hard. What constitutes an error varies from dataset to dataset. As a result, the standard practice is to manually explore and inspect the dataset using domain knowledge, iteratively formulating hypotheses about potential erroneous entries, correcting these errors, and repeating the process until the dataset is deemed clean \citep{ridzuan2019review}.

In recent years, LLMs have revolutionized various fields, including coding \citep{jiang2024survey},  writing \citep{li2024pre}, information search \citep{zhu2023large}. Even within the field of ML—the very domain from which LLMs originate—researchers are exploring their potential to automate tasks such as feature selection, model tuning, and optimization \citep{huang2024mlagentbench, kuken2024large}. Some studies suggest that these models have already surpassed median human performance for certain tasks \citep{chan2024mle}. 
However, most studies have focused on boosting ML model performance by optimizing model architecture and feature processing, rather than on enhancing the quality of the raw data.

Therefore, in this work, we shift the focus to data quality. Specifically, we ask: \textbf{If we keep the pre-processing and training pipeline fixed, can an LLM improve model performance purely by detecting and correcting errors in the data?} 

Our main contribution is the introduction of a simple framework for the systematic study of LLMs’ ability to interact with datasets to achieve a specific goal—namely, improving the quality of training data for ML downstream tasks and to provide an initial analysis of the LLMs' performance.

\section{Related Work}
\subsection{Data cleaning methods}
Data scientists have a wide array of tools available for detecting and correcting errors and inconsistencies. Classical approaches rely on predefined rules and statistical techniques to identify issues in syntax, semantics, and outliers. These methods might involve checking for violations of integrity constraints, identifying duplicate entries, or applying outlier detection algorithms \citep{fan2012foundations}. For error correction, problematic data is often replaced using external sources, basic statistical computations, or established integrity constraints \citep{ilyas2015trends}. However, these techniques require careful parameter tuning and rigid rule definitions, making the process both time-consuming and labor-intensive. To alleviate these challenges, several approaches have been developed: Raha \citep{mahdavi2019raha} and HoloDetect \citep{heidari2019holodetect} focus on error detection, while Baran \citep{mahdavi2020baran} addresses error correction. These methods leverage ML to reduce or even eliminate the reliance on manually defined rules. Furthermore, these methods tend to also have better overall performance than classical methods, however they struggle in case of rare or complex error types \citep{ni2023automatic}.
With the advent of LLMs, many works have begun exploring their utilisation for dataset cleaning motivated by the common sense reasoning of these models. \cite{narayan2022can} has explored using the GPT-3.5 architecture for various data wrangling tasks including error detection and data imputation, showing state of the art performance.  Other works, such as \cite{qi2024cleanagent} and \cite{li2024autodcworkflow}, also explore the use of LLMs for value standardization. Recently, IterClean \citep{ni2024iterclean} has claimed state of the art performance for entire detection-correction pipeline while using fewer supervised examples than previous methods.
However, while the results are impressive, current evaluation has been restricted only to established benchmarks of data cleaning, and it still unclear to which extent LLM can automate the entire data cleaning pipeline
\citep{majumder2024discoverybench}.

\subsection{Benchmarking LLM powered agents on data science tasks}
LLMs have been extensively studied for a variety of data science applications. The seminal work of \cite{lai2023ds} benchmarked LLMs for code generation in data science using thousands of problems specifically targeting data science tasks. Other studies have expanded the range of tasks and improved evaluation methodologies \citep{zhang2024benchmarking, huang2024code, galimzyanov2024drawing}.

Closely related to our work, several benchmarks move beyond simple question answering to enable function tool calling and multi-turn interactions. For example, \cite{liu2023rethinking} evaluates LLMs on question answering tasks that involve tabular data. Similarly, \cite{hu2024infiagent} proposes a framework for evaluating LLM-based agents on data analysis tasks—including data exploration, summary statistics, regression analysis, and feature engineering.
Moreover, \cite{majumder2024discoverybench} introduces a comprehensive benchmark that formalizes the entire multi-step process of data-driven discovery—from hypothesis formulation and verification using both real-world and synthetic tasks across diverse domains—to systematically assess and improve LLMs’ capabilities in automating scientific discovery. Likewise, \cite{chen2024scienceagentbench} presents a benchmark that assesses LLMs ability to autonomously generate executable Python programs for diverse, expert-validated data-driven scientific discovery tasks.
Finally, approaches such as \cite{huang2024mlagentbench} and \cite{chan2024mle}  benchmark LLMs on ML competitions with explicit target of improving performance on a test set. However, the focus in these works is on feature engineering and machine learning, rather than on improving the quality of the training data.

\section{Method}
\begin{figure}[t] 
    \centering
    \includegraphics[width=0.8\textwidth]{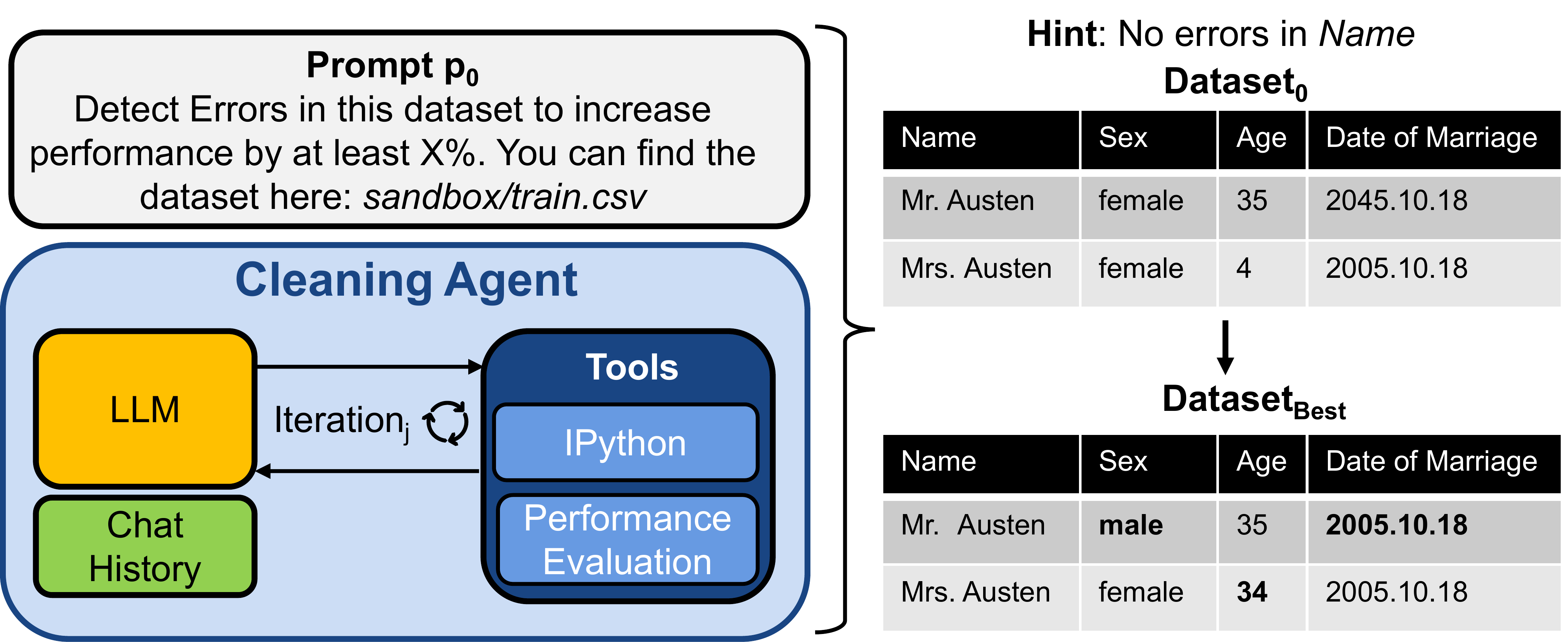} 
    \caption{
    We provide the model with the path to the dataset along with a prompt instructing it to identify errors so that performance on a held-out set increases by a given threshold. At each iteration \(j\), the LLM can send \textit{code} to IPython to execute and get back the \textit{sys.output} and/or send the \textit{path} of the modified dataset $\mathcal{D}_{\text{i}}$ to get a \textit{performance score}. The loop continues until the cumulative number of tokens used for the entire conversation reaches a pre-defined threshold. All the modified datasets $\mathcal{D}_{0...\text{i}}$ are stored and the dataset with the highest score is considered as $\mathcal{D}_{\text{Best}}$.}
    \label{fig:architecture} 
\end{figure}

Our goal is to evaluate how LLMs can effectively detect and correct errors in \textit{dirty} training datasets that negatively impact model performance on a held-out dataset. In our setup, the LLM has access to two tools. The first is the interactive Python shell IPython \citep{PER-GRA:2007}, which executes Python code. Using this tool, the LLM can inspect and modify the dirty dataset used to train a ML model. The second is the performance evaluation. It can iteratively submit a \textit{modified} version of the training dataset to an evaluation pipeline, which returns the performance score of the model trained on the this new data and evaluated on a \textit{clean} held-out evaluation set. The pre-processing, training, and evaluation pipelines are kept fixed and non-modifiable by the LLM, ensuring that the LLM can influence model performance solely by altering the training dataset. Figure \ref{fig:architecture} shows an overview of our approach. In the following sections, we outline the procedure for generating dirty versions of the datasets and describe how the LLM interacts with them.

\subsection{Dataset creation}\label{sec:datasets_creation}
While some datasets in the literature \citep{rekatsinas2017holoclean,mahdavi2019raha} provide both clean and dirty versions, the data quality issues in the corrupted versions do not significantly impair the performance of downstream tasks \citep{ni2023automatic}.

To address this limitation, we create our own clean and dirty dataset versions as follows. First, we identify popular datasets from \href{https://www.kaggle.com/}{Kaggle}. Each dataset, denoted as $\mathcal{D}$, is divided into a training set $\mathcal{D}_{\text{TrainClean}}$ and a testing set $\mathcal{D}_{\text{TestClean}}$, which serves as a held-out evaluation set that remains non-modifiable and unseen by both the model and the LLM. We train a model on $\mathcal{D}_{\text{TrainClean}}$ and evaluate its performance on $\mathcal{D}_{\text{TestClean}}$, establishing a baseline performance, $P_{\text{Clean}}$.
Next, we introduce three systematic errors in $\mathcal{D}_{\text{TrainClean}}$, resulting in $\mathcal{D}_{\text{TrainDirty}}$. These errors are selected from the categories defined below (Numerical Shift, NaN Corruption, and Categorical Shift).
We re-train the model on $\mathcal{D}_{\text{TrainDirty}}$ and again evaluate its performance on $\mathcal{D}_{\text{TestClean}}$, yielding a performance score of $P_{\text{Dirty}}$.  

The difference between $P_{\text{Clean}}$ and $P_{\text{Dirty}}$ represents the maximum potential performance improvement achievable by correcting the introduced systematic errors. However, $P_{\text{Clean}}$ is not strictly an upper bound on performance, as other inherent issues within the Kaggle datasets may exist that, if addressed, could lead to even higher performance.

We define the following three categories of systematic errors introduced in $\mathcal{D}_{\text{TrainClean}}$:
\begin{itemize}
    \item \textit{Numerical Shift}: Alters numerical values to introduce a distribution shift in a subset of the dataset (e.g., increasing all age entries by $10$ years or restricting them to a range between $0$ and $10$).
    \item \textit{NaN Corruption}: Replaces a fraction of well-defined values with NaN in a subset of the dataset, following a missing at random corruption strategy (e.g., corrupting with NaN the values of a specific column where a specific condition happens).
     \item \textit{Categorical Shift}: Shifts the distribution of categorical feature values within a subset of the dataset, leading to statistical or contextual inconsistencies (e.g., changing the category of a feature to one that is uncommon or contextually inappropriate for that subset).
\end{itemize}

In order to create a fair evaluation scenario, we also ensure that the introduced errors meet the following criteria: 
\begin{enumerate}
    \item They are detectable through statistical analysis, contextual understanding of the dataset, or a combination of both; and
    \item The resulting decrease in model performance due to dataset corruption can be mitigated through dataset manipulations that do not require access to the original \textit{clean} dataset.
\end{enumerate}

Additionally, to gain preliminary insights into how humans compare to LLMs on this task, we asked two data scientists to solve the same task as the LLMs on each dataset within a time limit of one hour and recorded their peformance improvement over $P_{Dirty}$. This also offers an initial perspective on the perceived difficulty of detecting and correcting errors in our datasets. Details on the datasets and the introduced errors can be found in the Appendix \ref{sec:datasets_details}.


\subsection{Cleaning Agent}
We provide an initial prompt $P_{0}$ to the LLM, which includes clear instructions for the task, examples of inconsistencies, the path to $\mathcal{D}_{\text{TrainDirty}}$, explanations of the available tools, and guidance on how to solve the task effectively. The full prompt is provided in the Appendix \ref{sec:initial_prompt}. The LLM has access to the following two tools:

\begin{itemize}

\item \textbf{Performance Evaluation (}\textbf{Input}: \textit{DatasetPath}, \textbf{Outptut}: \textit{PerformanceScore}\textbf{)}: This tool submits the dataset modified by the LLM and evaluates it. It takes as input the path to $\mathcal{D}_{\text{TrainModified}_{i}}$, performs pre-defined pre-processing steps, trains the learning algorithm on it, and returns the model’s performance score on a classification task using $\mathcal{D}_{\text{TestClean}}$. Note that the LLM cannot modify the evaluation pipeline or add new columns, though it is permitted to drop columns or rows if necessary.


\item \textbf{IPython (Input: }\textit{PythonCode},\textbf{ Output: }\textit{StandardOutput}): Executes the code provided as input, using the interactive IPython shell and returns the standard output produced during execution. This includes the output of print statements, the contents of defined variables, and any errors encountered. The shell preserves a persistent session state across executions, so variables and outputs from earlier runs remain available in subsequent executions.
\end{itemize}

The LLM’s interactions with the dataset begin after the initial prompt $P_{0}$ is provided. An iteration $j$ starts when the LLM generates text, and in each iteration, the LLM may invoke one or both tools. The responses from these tools, along with the previous prompt and the LLM’s output from the current iteration, are appended to form the prompt for the next iteration (i.e., $p_{j+1} = p_{j} + output_{j} + ToolResponse_{j}$). If the LLM wants to submit a modified version of the training dataset $\mathcal{D}_{\text{TrainModified}_{i}}$, it must save the dataset and provide its path to the \textit{Performance Evaluation} tool, which then returns the score for that submission. The loop terminates when the cumulative number of tokens used (input plus output, summed over all iterations) reaches a predefined threshold, and the dataset submission with the highest score is considered the LLM’s best result.




\section{Experiments and results}
\subsection{Setup}
We consider three popular datasets from Kaggle: \href{https://www.kaggle.com/datasets/yasserh/titanic-dataset}{Titantic}, \href{https://www.kaggle.com/datasets/scibearia/meat-consumption-per-capita}{Meat Consumptions}, \href{https://www.kaggle.com/datasets/mojtaba142/hotel-booking}{Hotel Bookings}. A classification task is defined for each dataset. We perform minimal pre-processing and ensure the classification tasks remain challenging by removing columns that contain information which would make the tasks trivial. A standard train/test split is performed, and, as outlined in Section~\ref{sec:datasets_creation}, three different types of systematic errors are introduced into the training data. Each of these errors requires investigating rows, columns, or both within the dataset. Further details on the errors are provided in the Appendix~\ref{sec:datasets_details}. 
We conduct experiments using the following models with the function calling feature: \texttt{gpt-4o}, \texttt{o3-mini-2025-01-31}, \texttt{claude-3-5-sonnet-20241022}, and \texttt{gemini-2.0-flash-exp}. Each experiment is run with a limit of 200k tokens and repeated six times to ensure consistency. Our code structure is based on Swarm from \cite{OpenAI_Swarm} due to its lightweight design and flexibility. Table \ref{tab:datasets_exploration} provides an overview of the different datasets including the data scientist performance and the type of errors introduced.


\begin{table}[ht]
\centering
\renewcommand{\arraystretch}{1.1} 
\begin{tabular}{l l l l l}
\toprule
\textbf{Dataset} & \textbf{Shape} & \textbf{Data Scientist Perf.} & \textbf{Error Types} & \\
& & \textbf{Improvement [1h]} & & \\ 
\midrule
Titanic           & (408, 12)    & 6.2\%  \textit{(easy)} &  \parbox[t]{0.17\textwidth}{Numerical Shift  \\ Categorical Shift}   \\ 
Meat Consumption  & (9160, 10)   & 2.5\%  \textit{(medium)} & Numerical Shift    \\ 
Hotel Bookings    & (100000, 28) & 0\%  \textit{(hard)} & \parbox[t]{0.17\textwidth}{Numerical Shift \\ Nan Corruption \\ Categorical Shift}       & \\ 
\bottomrule
\end{tabular}
\caption{Comparison of the three Kaggle datasets in terms of shape, the performance improvement achieved by humans in one hour and error types.
The perceived difficulty of each dataset (\textit{easy}, \textit{medium}, \textit{hard}) as reported by the human participants is also provided. }
\label{tab:datasets_exploration}
\end{table}



\subsection{Results}
In the following sections, we first present our quantitative results alongside a qualitative discussion to interprets them. Next, we analyze how cumulative token consumption and the strength of provided hints affect performance improvements. Finally, we describe recurring issues observed in the model interactions.

\subsubsection{How effective are LLMs at improving model performance by correcting errors?}\label{sec:effective_performance}

\begin{figure}[htbp]
    \centering
    \includegraphics[width=0.8\textwidth]{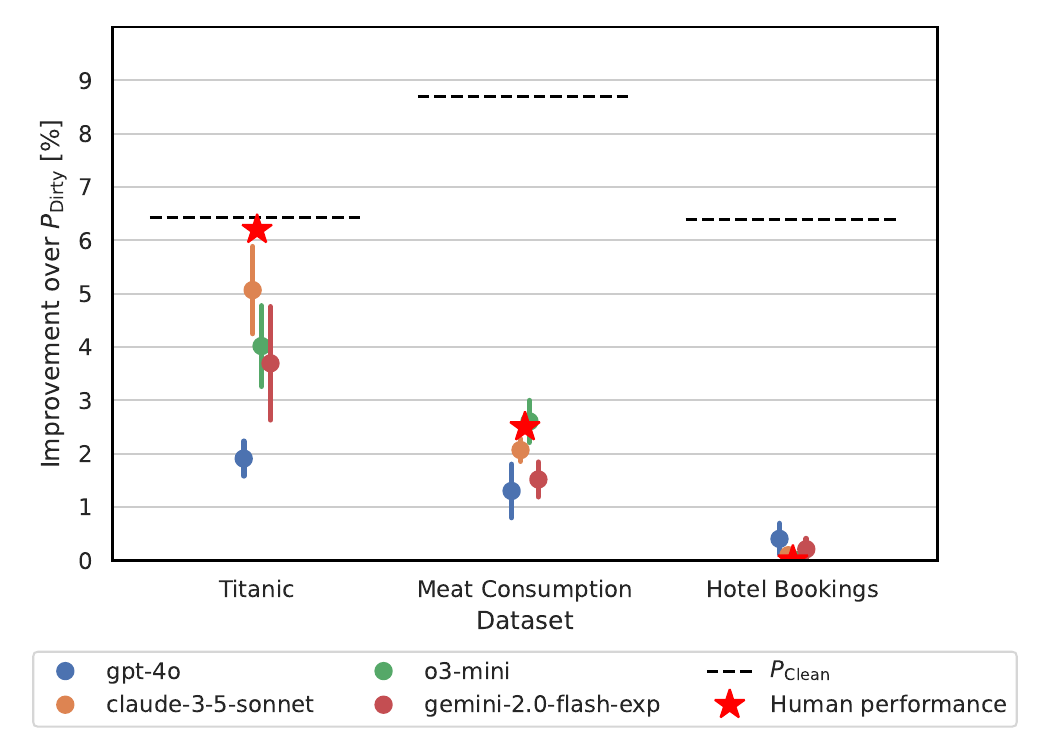} 
    \caption{Performance improvement over $P_{\text{Dirty}}$ for the four models and three datasets.}
    \label{fig:no_hint} 
\end{figure}

Figure \ref{fig:no_hint} shows that 
no model reaches the maximum potential performance improvement, i.e., achieving $P_{\text{Clean}}$. A significant gap between the achieved performance improvement and $P_{\text{Clean}}$ is noticeable for both the \textit{Meat Consumption} and \textit{Hotel Bookings} datasets, highlighting their increased difficulty. Specifically, for \textit{Hotel Bookings}, all models achieve less than $1\%$ improvement, which emphasizes the challenging errors introduced in this dataset. Interestingly, the perceived difficulty of the datasets by humans is also reflected in the LLMs' performance, with the \textit{Titanic} dataset being the easiest, followed by \textit{Meat Consumption} and \textit{Hotel Bookings}. Overall, while \textit{o3-mini} and \textit{claude-3-5-sonnet} show a slight edge over the others for the \textit{Titanic} and \textit{Meat Consumption} datasets, no model clearly outperforms the rest.

To better understand where the models face challenges, we manually inspect the conversation traces from the runs. Listing~\ref{lst:listing_1} shows the code generated in the first response by \textit{o3-mini} during a sample interaction on the \textit{Meat Consumption} dataset. We observe that the models begin their analysis by exploring the data using pandas functions such as \verb|.describe()|, \verb|.info()|, and \verb|.head()|, along with performing simple sanity checks on individual entries. However, models rarely investigate more complex relationships within the data.


\begin{lstlisting}[language=Python, caption=Python code generated in a first response by o3-mini on the Meat Consumption dataset., label={lst:listing_1}]
print(df.head())
print(df.describe(include='all'))

columns_to_check = ['Poultry', 'Beef', 'Sheep and goat', 'Pork', 'Other meats', 'Fish and seafood']
print('\nNegative value counts:')
for col in columns_to_check:
    if col in df.columns:
        neg_count = (df[col] < 0).sum()
        print(f'{col}: {neg_count}')
    else:
        print(f'{col} not found in dataset')

print('\nMax values in consumption columns:')
for col in columns_to_check:
    if col in df.columns:
        max_val = df[col].max()
        print(f'{col}: {max_val}')

print('\nMissing value counts:')
print(df.isnull().sum())
\end{lstlisting}

As a result, errors that require the analysis of multiple values in the dataset—such as identifying distribution shifts, trends, or biases— go undetected. Even when errors are identified, limited exploration across different values hinders effective error mitigation. For example, no model was able to detect the numerical shift introduced in the \textit{Hotel Bookings} dataset, where 10 was added to all values in the \texttt{lead\_time} column only for the year 2016. This represents a simple bias that the models failed to identify.  Listing~\ref{lst:listing_2} shows an example from the \textit{Meat Consumption} dataset, where \textit{claude-3-5} correctly detects that some landlocked countries, such as Afghanistan and Nepal, have abnormally high fish consumption. However, instead of applying a statistically meaningful approach, such as using quantiles, it applies an arbitrary scaling factor. 


\begin{lstlisting}[language=Python, 
caption={Successful error detection, but arbitrary error mitigation by claude-3-5 in the Meat Consumption dataset.}, 
label={lst:listing_2}]
landlocked_countries = ['Afghanistan', 'Mongolia', 'Nepal', 'Bhutan', 'Laos', 'Uganda']
for country in landlocked_countries:
    mask = (df_cleaned['Entity'] == country) & (df_cleaned['Fish and seafood'] > 15)
    df_cleaned.loc[mask, 'Fish and seafood'] = df_cleaned.loc[mask, 'Fish and seafood'] * 0.3
\end{lstlisting}

We observe the same behavior for categorical variables. For example, in the \textit{Hotel Bookings} dataset, the NaN corruption involved replacing $70\%$ of entries with NaN where \verb|country == 'PRT'| in the years 2016 and 2017. All models, except for \textit{gpt-4o} in a single run, fail to address this error correctly. Instead of recognizing that \verb|PRT| is the most frequent country based on the available data from 2015, they replace the missing values with a new entry labeled \textit{Other}, as shown in Listing~\ref{lst:listing_3}. 


\begin{lstlisting}[language=Python, caption=Arbitrary error mitigation by o3-mini in the Hotel Bookings dataset., label={lst:listing_3}]
# Fix negative adr: replace negative adr with median of non-negative adr
adr_median = df.loc[df['adr'] >= 0, 'adr'].median()
df.loc[df['adr'] < 0, 'adr'] = adr_median
# Fill missing values
# For country, fill missing with 'Other'
df['country'] = df['country'].fillna('Other')
\end{lstlisting}

\subsubsection{How many tokens are needed to achieve the best performance?}\label{sec:token_impact}





\begin{figure}[htbp] 
    \centering
    \includegraphics[width=1\textwidth]{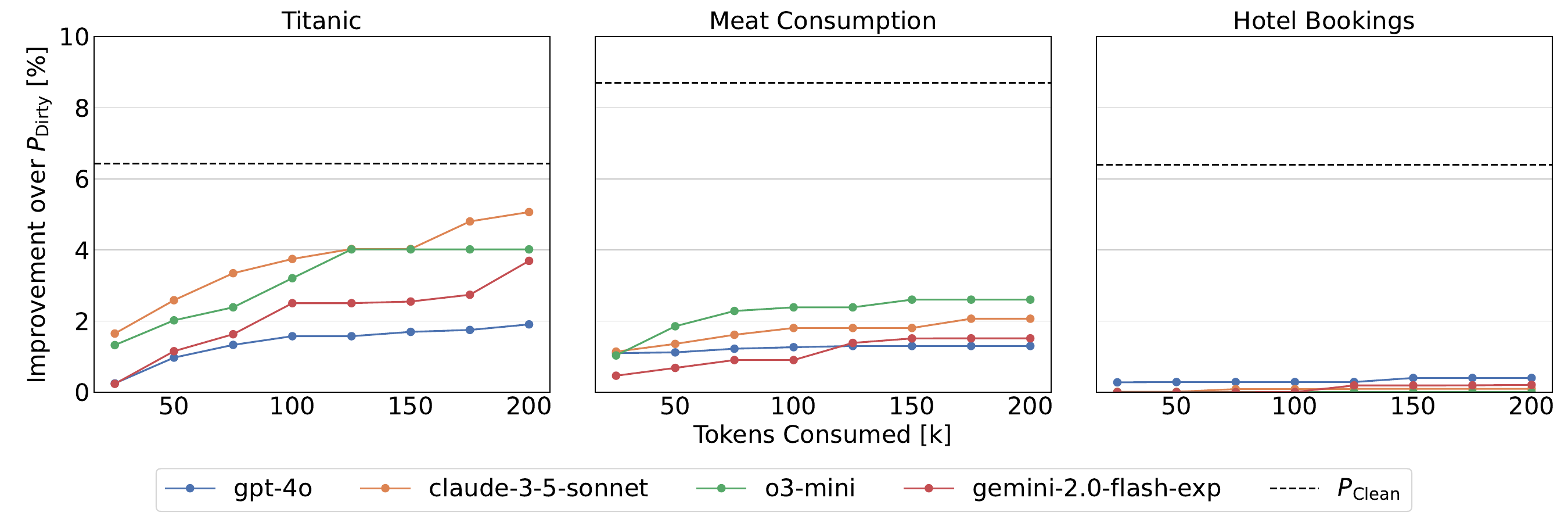} 
    \caption{Performance improvement for different \textit{Cumulative tokens} thresholds from 25k to 200k}
    \label{fig:time_thinking} 
\end{figure}

Figure~\ref{fig:time_thinking} shows the cumulative tokens consumed, ranging from 25k up to a maximum of 200k, in relation to the best achieved improvement in performance. Across all models and datasets, a logarithmic trend is observed, indicating diminishing returns as token consumption increases. Given that \textit{gemini-2.0-flash-exp} supports a context window of up to 1M tokens, we conduct an additional experiment by increasing the cumulative token consumption limit to 2M for this model. Table~\ref{tab:performance_table} shows that, for the \textit{Titanic} and \textit{Hotel Bookings} datasets, performance increases by $5.19\%$ and $1.82\%$, respectively, compared to the results with a $200$k token limit. This suggests that the \textit{gemini-2.0-flash-exp} benefits significantly from the larger context window, possibly allowing for more thorough exploration and error correction. 




\begin{table}[H]
\centering
\begin{tabular}{lccc}
\toprule
\textbf{Dataset} & \textbf{200k Tokens} & \textbf{2M Tokens} \\ \midrule
Titanic          & 3.69\%               & 8.88\%             \\
Meat Consumption & 1.51\%               & 1.79\%             \\
Hotel Bookings   & 0.08\%               & 1.90\%             \\ \bottomrule
\end{tabular}
\caption{Performance improvements of \textit{gemini-2.0-flash-exp} with 200k and 2 million cumulative token limits across all datasets.}
\label{tab:performance_table}
\end{table}

\subsubsection{Do Hints help?}\label{sec:hints_help}

\begin{figure}[htbp] 
    \centering
    \includegraphics[width=0.99\textwidth]{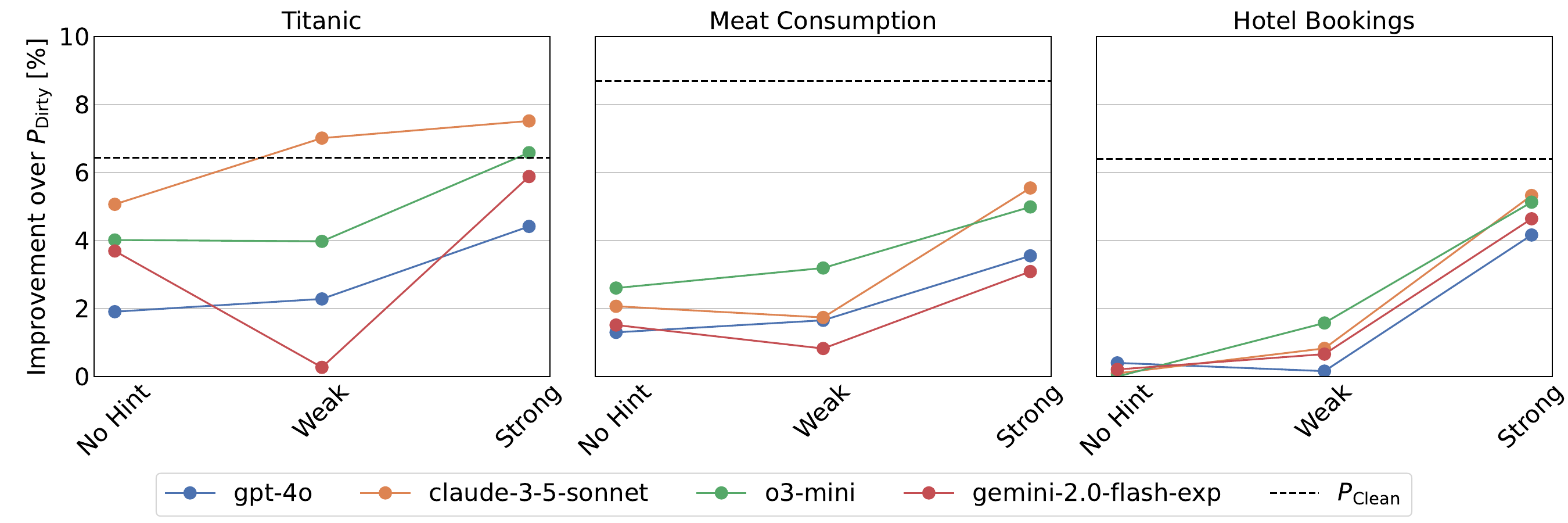} 
    \caption{Impact of providing \textit{no hint}, a \textit{weak} hint, and a \textit{strong} hint on the performance improvement for all models and datasets.}
    \label{fig:after_hint} 
\end{figure}

Figure \ref{fig:after_hint} shows how performance improvement across all models and datasets changes depending on how much additional context is provided to the model in the initial prompt $P_{0}$. The models are evaluated under three conditions: no hint, a \textit{weak} hint, or a \textit{strong} hint. A weak hint provides partial information about the location of the error in the dataset, while a strong hint offers complete information about the error's location and partial guidance on how to correct it. Examples of these hints for all datasets are provided in the Appendix \ref{sec:datasets_details}.

Overall, a clear trend emerges: providing hints generally enhances performance, as the models can leverage the extra contextual information to improve error detection and correction. However, there are exceptions. We noticed that, on some occasions, the weak hint causes the model to apply an inappropriate error correction strategy to a column—one that actually decreases performance—whereas, with no hint, that column would remain unchanged.

\subsubsection{Additional Remarks}\label{sec:remarks}
Beyond the points mentioned earlier, our analysis of the conversation traces revealed several recurring issues common to all models. In particular, we observed:


\begin{itemize}
    \item Models frequently fail to submit a valid dataset. As shown in Table~\ref{table:failure}, a significant percentage of submissions contain errors. Major reasons include providing a file path that does not exist and submitting datasets with extra columns.
    
    \item Models tends to apply a brute-force approach by repeatedly submitting datasets without meaningful analysis. Table~\ref{table:calls_to_python} shows the fraction of dataset submissions relative to the generated code, highlighting that in the majority of iterations, the models prioritize submitting datasets over generating code for meaningful analysis.
    
    \item Models (with the exception of gpt-4o) seems to ignore that the state is preserved in IPython. Specifically, in each tool call, the code is generated from scratch rather than building on outputs from previous executions. This hinders in-depth exploration of the dataset across multiple iterations.

\end{itemize}


\begin{table}[H]
\centering
\begin{tabular}{l l l l l}
\toprule
\textbf{Model} & \textbf{Column} & \textbf{Dataset} & \textbf{Other [\%]} & \textbf{Total} \\  & \textbf{Violation [\%]} & \textbf{not found [\%]} & & \textbf{Failures  [\%]} \\
\midrule
claude-3-5-sonnet    & 8.84  & 0.00  & 4.69  & 13.53 \\
gemini-2.0-flash-exp & 5.05  & 4.32  & 3.73  & 13.10 \\
gpt-4o               & 5.87  & 0.00  & 2.07  & 7.94  \\
o3-mini              & 0.00  & 5.40  & 5.47  & 10.87 \\ 
\bottomrule
\end{tabular}
\caption{Percentage of invalid submissions for each model. \textit{Column Violation} indicates that new columns were added to the \textit{modified} dataset, violating the rules described in the initial prompt $P_{0}$. \textit{Dataset not found} means the dataset path provided by the LLM to the evaluation function does not exist. \textit{Other} refers to additional violation errors.}
\label{table:failure}
\end{table}

\begin{table}[H]
\centering
\begin{tabular}{l c}
\toprule
\textbf{Model} & \textbf{Dataset related submissions} \\ 
& \textbf{over total tool calls [\%]}  \\ 
\midrule
claude-3-5-sonnet    & 61.86\% \\ 
gemini-2.0-flash-exp & 63\%    \\ 
gpt-4o               & 29\%    \\ 
o3-mini              & 84.70\% \\ 
\bottomrule
\end{tabular}
\caption{Percentage of calls to the IPython tool related to the creation of a new submission dataset compared to the total number of calls.}
\label{table:calls_to_python}
\end{table}


\section{Discussion}
\subsection{Conclusive Remarks}
We present a simple strategy to benchmark LLMs in one of the most crucial yet often overlooked activities of data scientists: cleaning data prior to model training. We propose a pipeline in which the LLM has access to IPython to programmatically modify a training dataset that has been corrupted with errors. The processes of training the ML model and performing any feature engineering are kept fixed and are not modifiable by the LLM. The LLM can \textit{iteratively} explore the dataset to detect and correct errors, receiving feedback based on the model’s performance when trained on its submitted modified datasets and evaluated on a clean, held-out test set. Our goal is to explore the strength and limitations of the current state-of-the-art LLMs on this cleaning task across different datasets. Our results show that while LLMs can detect and mitigate errors in two of the three datasets considered, none of them achieve the maximum potential performance improvement achievable by fully correcting the introduced systematic errors. Overall, providing the LLM with hints about the errors leads to improved performance, indicating that the models can effectively process contextual information to solve the given task. An analysis of the conversation traces reveals that LLMs are able to identify and correct errors that involve investigating \textit{single} values or \textit{individual} rows. However, they struggle with errors spanning multiple rows, such as distribution shifts, trends, or biases.


\subsection{Limitations and Future Work}
Our work can be extended from multiple angles. First, the current setup enforces text-based communication between the LLM and the tools. This could be extended to allow the LLM to receive visual artifacts, such as plots generated from code, which are often essential during the data exploration phase. Second, our current approach involves manually introducing and validating errors, which is labor-intensive and limits the scalability of the benchmark to a wider range of datasets. Moreover, since only the authors were involved in creating errors for our three datasets, there is a potential risk of limited diversity and biases influencing our findings. Future work should focus on automating error generation to ensure a diverse and realistic set of errors across various datasets, thereby mitigating potential biases. Third, some of the insights in this work were derived from manually inspecting the raw messages exchanged at each iteration between the LLM and the tools. Future research should focus on automating this process to enable quantifiable measurements and address the following questions: 1) What types of failures affect the LLM, such as reasoning flaws, ineffective exploration, or other factors? and 2) At which stages these failures occur most frequently—whether during error detection or correction. Finally, the impact of different prompting approaches should be explored. Specifically, future research should investigate how varying prompts influence the LLM’s performance, as well as its error detection and correction strategies. We plan to extend this work by addressing these limitations.

\bibliography{iclr2025_conference}

\begin{thebibliography}{30}
\providecommand{\natexlab}[1]{#1}
\providecommand{\url}[1]{\texttt{#1}}
\expandafter\ifx\csname urlstyle\endcsname\relax
  \providecommand{\doi}[1]{doi: #1}\else
  \providecommand{\doi}{doi: \begingroup \urlstyle{rm}\Url}\fi

\bibitem[Chan et~al.(2024)Chan, Chowdhury, Jaffe, Aung, Sherburn, Mays, Starace, Liu, Maksin, Patwardhan, et~al.]{chan2024mle}
Jun~Shern Chan, Neil Chowdhury, Oliver Jaffe, James Aung, Dane Sherburn, Evan Mays, Giulio Starace, Kevin Liu, Leon Maksin, Tejal Patwardhan, et~al.
\newblock Mle-bench: Evaluating machine learning agents on machine learning engineering.
\newblock \emph{arXiv preprint arXiv:2410.07095}, 2024.

\bibitem[Chen et~al.(2024)Chen, Chen, Ning, Zhang, Wang, Yu, Li, Liao, Wei, Lu, et~al.]{chen2024scienceagentbench}
Ziru Chen, Shijie Chen, Yuting Ning, Qianheng Zhang, Boshi Wang, Botao Yu, Yifei Li, Zeyi Liao, Chen Wei, Zitong Lu, et~al.
\newblock Scienceagentbench: Toward rigorous assessment of language agents for data-driven scientific discovery.
\newblock \emph{arXiv preprint arXiv:2410.05080}, 2024.

\bibitem[Fan \& Geerts(2012)Fan and Geerts]{fan2012foundations}
Wenfei Fan and Floris Geerts.
\newblock \emph{Foundations of data quality management}.
\newblock Morgan \& Claypool Publishers, 2012.

\bibitem[Galimzyanov et~al.(2024)Galimzyanov, Titov, Golubev, and Bogomolov]{galimzyanov2024drawing}
Timur Galimzyanov, Sergey Titov, Yaroslav Golubev, and Egor Bogomolov.
\newblock Drawing pandas: A benchmark for llms in generating plotting code.
\newblock \emph{arXiv preprint arXiv:2412.02764}, 2024.

\bibitem[Heidari et~al.(2019)Heidari, McGrath, Ilyas, and Rekatsinas]{heidari2019holodetect}
Alireza Heidari, Joshua McGrath, Ihab~F Ilyas, and Theodoros Rekatsinas.
\newblock Holodetect: Few-shot learning for error detection.
\newblock In \emph{Proceedings of the 2019 International Conference on Management of Data}, pp.\  829--846, 2019.

\bibitem[Hu et~al.(2024)Hu, Zhao, Wei, Chai, Ma, Wang, Wang, Su, Xu, Zhu, et~al.]{hu2024infiagent}
Xueyu Hu, Ziyu Zhao, Shuang Wei, Ziwei Chai, Qianli Ma, Guoyin Wang, Xuwu Wang, Jing Su, Jingjing Xu, Ming Zhu, et~al.
\newblock Infiagent-dabench: Evaluating agents on data analysis tasks.
\newblock \emph{arXiv preprint arXiv:2401.05507}, 2024.

\bibitem[Huang et~al.(2024{\natexlab{a}})Huang, Vora, Liang, and Leskovec]{huang2024mlagentbench}
Qian Huang, Jian Vora, Percy Liang, and Jure Leskovec.
\newblock Mlagentbench: Evaluating language agents on machine learning experimentation.
\newblock In \emph{Forty-first International Conference on Machine Learning}, 2024{\natexlab{a}}.

\bibitem[Huang et~al.(2024{\natexlab{b}})Huang, Luo, Yu, Zhang, Lei, Wei, He, Huang, Liu, Zhao, et~al.]{huang2024code}
Yiming Huang, Jianwen Luo, Yan Yu, Yitong Zhang, Fangyu Lei, Yifan Wei, Shizhu He, Lifu Huang, Xiao Liu, Jun Zhao, et~al.
\newblock Da-code: Agent data science code generation benchmark for large language models.
\newblock \emph{arXiv preprint arXiv:2410.07331}, 2024{\natexlab{b}}.

\bibitem[Ilyas et~al.(2015)Ilyas, Chu, et~al.]{ilyas2015trends}
Ihab~F Ilyas, Xu~Chu, et~al.
\newblock Trends in cleaning relational data: Consistency and deduplication.
\newblock \emph{Foundations and Trends{\textregistered} in Databases}, 5\penalty0 (4):\penalty0 281--393, 2015.

\bibitem[J{\"a}ger \& Biessmann(2024)J{\"a}ger and Biessmann]{jager2024data}
Sebastian J{\"a}ger and Felix Biessmann.
\newblock From data imputation to data cleaning—automated cleaning of tabular data improves downstream predictive performance.
\newblock In \emph{International Conference on Artificial Intelligence and Statistics}, pp.\  3394--3402. PMLR, 2024.

\bibitem[Jiang et~al.(2024)Jiang, Wang, Shen, Kim, and Kim]{jiang2024survey}
Juyong Jiang, Fan Wang, Jiasi Shen, Sungju Kim, and Sunghun Kim.
\newblock A survey on large language models for code generation.
\newblock \emph{arXiv preprint arXiv:2406.00515}, 2024.

\bibitem[K{\"u}ken et~al.(2024)K{\"u}ken, Purucker, and Hutter]{kuken2024large}
Jaris K{\"u}ken, Lennart Purucker, and Frank Hutter.
\newblock Large language models engineer too many simple features for tabular data.
\newblock \emph{arXiv preprint arXiv:2410.17787}, 2024.

\bibitem[Lai et~al.(2023)Lai, Li, Wang, Zhang, Zhong, Zettlemoyer, Yih, Fried, Wang, and Yu]{lai2023ds}
Yuhang Lai, Chengxi Li, Yiming Wang, Tianyi Zhang, Ruiqi Zhong, Luke Zettlemoyer, Wen-tau Yih, Daniel Fried, Sida Wang, and Tao Yu.
\newblock Ds-1000: A natural and reliable benchmark for data science code generation.
\newblock In \emph{International Conference on Machine Learning}, pp.\  18319--18345. PMLR, 2023.

\bibitem[Li et~al.(2024{\natexlab{a}})Li, Tang, Zhao, Nie, and Wen]{li2024pre}
Junyi Li, Tianyi Tang, Wayne~Xin Zhao, Jian-Yun Nie, and Ji-Rong Wen.
\newblock Pre-trained language models for text generation: A survey.
\newblock \emph{ACM Computing Surveys}, 56\penalty0 (9):\penalty0 1--39, 2024{\natexlab{a}}.

\bibitem[Li et~al.(2024{\natexlab{b}})Li, Fang, and Torvik]{li2024autodcworkflow}
Lan Li, Liri Fang, and Vetle~I Torvik.
\newblock Autodcworkflow: Llm-based data cleaning workflow auto-generation and benchmark.
\newblock \emph{arXiv preprint arXiv:2412.06724}, 2024{\natexlab{b}}.

\bibitem[Liu et~al.(2023)Liu, Wang, and Chen]{liu2023rethinking}
Tianyang Liu, Fei Wang, and Muhao Chen.
\newblock Rethinking tabular data understanding with large language models.
\newblock \emph{arXiv preprint arXiv:2312.16702}, 2023.

\bibitem[Mahdavi \& Abedjan(2020)Mahdavi and Abedjan]{mahdavi2020baran}
Mohammad Mahdavi and Ziawasch Abedjan.
\newblock Baran: Effective error correction via a unified context representation and transfer learning.
\newblock \emph{Proceedings of the VLDB Endowment}, 13\penalty0 (12):\penalty0 1948--1961, 2020.

\bibitem[Mahdavi et~al.(2019)Mahdavi, Abedjan, Castro~Fernandez, Madden, Ouzzani, Stonebraker, and Tang]{mahdavi2019raha}
Mohammad Mahdavi, Ziawasch Abedjan, Raul Castro~Fernandez, Samuel Madden, Mourad Ouzzani, Michael Stonebraker, and Nan Tang.
\newblock Raha: A configuration-free error detection system.
\newblock In \emph{Proceedings of the 2019 International Conference on Management of Data}, pp.\  865--882, 2019.

\bibitem[Majumder et~al.(2024)Majumder, Surana, Agarwal, Mishra, Meena, Prakhar, Vora, Khot, Sabharwal, and Clark]{majumder2024discoverybench}
Bodhisattwa~Prasad Majumder, Harshit Surana, Dhruv Agarwal, Bhavana~Dalvi Mishra, Abhijeetsingh Meena, Aryan Prakhar, Tirth Vora, Tushar Khot, Ashish Sabharwal, and Peter Clark.
\newblock Discoverybench: Towards data-driven discovery with large language models.
\newblock \emph{arXiv preprint arXiv:2407.01725}, 2024.

\bibitem[Narayan et~al.(2022)Narayan, Chami, Orr, Arora, and R{\'e}]{narayan2022can}
Avanika Narayan, Ines Chami, Laurel Orr, Simran Arora, and Christopher R{\'e}.
\newblock Can foundation models wrangle your data?
\newblock \emph{arXiv preprint arXiv:2205.09911}, 2022.

\bibitem[Ni et~al.(2023)Ni, Miao, Zhao, Wu, and Yin]{ni2023automatic}
Wei Ni, Xiaoye Miao, Xiangyu Zhao, Yangyang Wu, and Jianwei Yin.
\newblock Automatic data repair: Are we ready to deploy?
\newblock \emph{arXiv preprint arXiv:2310.00711}, 2023.

\bibitem[Ni et~al.(2024)Ni, Zhang, Miao, Zhao, Wu, and Yin]{ni2024iterclean}
Wei Ni, Kaihang Zhang, Xiaoye Miao, Xiangyu Zhao, Yangyang Wu, and Jianwei Yin.
\newblock Iterclean: An iterative data cleaning framework with large language models.
\newblock In \emph{Proceedings of the ACM Turing Award Celebration Conference-China 2024}, pp.\  100--105, 2024.

\bibitem[OpenAI(2024)]{OpenAI_Swarm}
OpenAI.
\newblock Swarm, 2024.
\newblock URL \url{https://github.com/openai/swarm}.

\bibitem[P\'erez \& Granger(2007)P\'erez and Granger]{PER-GRA:2007}
Fernando P\'erez and Brian~E. Granger.
\newblock {IP}ython: a system for interactive scientific computing.
\newblock \emph{Computing in Science and Engineering}, 9\penalty0 (3):\penalty0 21--29, May 2007.
\newblock ISSN 1521-9615.
\newblock \doi{10.1109/MCSE.2007.53}.
\newblock URL \url{https://ipython.org}.

\bibitem[Qi \& Wang(2024)Qi and Wang]{qi2024cleanagent}
Danrui Qi and Jiannan Wang.
\newblock Cleanagent: Automating data standardization with llm-based agents.
\newblock \emph{arXiv preprint arXiv:2403.08291}, 2024.

\bibitem[Rekatsinas et~al.(2017)Rekatsinas, Chu, Ilyas, and R{\'e}]{rekatsinas2017holoclean}
Theodoros Rekatsinas, Xu~Chu, Ihab~F Ilyas, and Christopher R{\'e}.
\newblock Holoclean: Holistic data repairs with probabilistic inference.
\newblock \emph{arXiv preprint arXiv:1702.00820}, 2017.

\bibitem[Ridzuan \& Zainon(2019)Ridzuan and Zainon]{ridzuan2019review}
Fakhitah Ridzuan and Wan Mohd Nazmee~Wan Zainon.
\newblock A review on data cleansing methods for big data.
\newblock \emph{Procedia Computer Science}, 161:\penalty0 731--738, 2019.

\bibitem[Sambasivan et~al.(2021)Sambasivan, Kapania, Highfill, Akrong, Paritosh, and Aroyo]{sambasivan2021everyone}
Nithya Sambasivan, Shivani Kapania, Hannah Highfill, Diana Akrong, Praveen Paritosh, and Lora~M Aroyo.
\newblock “everyone wants to do the model work, not the data work”: Data cascades in high-stakes ai.
\newblock In \emph{proceedings of the 2021 CHI Conference on Human Factors in Computing Systems}, pp.\  1--15, 2021.

\bibitem[Zhang et~al.(2024)Zhang, Jiang, Han, Chen, Yang, and Ren]{zhang2024benchmarking}
Yuge Zhang, Qiyang Jiang, Xingyu Han, Nan Chen, Yuqing Yang, and Kan Ren.
\newblock Benchmarking data science agents.
\newblock \emph{arXiv preprint arXiv:2402.17168}, 2024.

\bibitem[Zhu et~al.(2023)Zhu, Yuan, Wang, Liu, Liu, Deng, Chen, Liu, Dou, and Wen]{zhu2023large}
Yutao Zhu, Huaying Yuan, Shuting Wang, Jiongnan Liu, Wenhan Liu, Chenlong Deng, Haonan Chen, Zheng Liu, Zhicheng Dou, and Ji-Rong Wen.
\newblock Large language models for information retrieval: A survey.
\newblock \emph{arXiv preprint arXiv:2308.07107}, 2023.

\end{thebibliography}
\bibliographystyle{iclr2025_conference}

\newpage
\appendix
\section{Appendix}


\subsection{Datasets}\label{sec:datasets_details}
In the following, we briefly describe the three errors introduced in the \href{https://www.kaggle.com/datasets/yasserh/titanic-dataset}{Titanic}, \href{https://www.kaggle.com/datasets/scibearia/meat-consumption-per-capita}{Meat Consumption}, and \href{https://www.kaggle.com/datasets/mojtaba142/hotel-booking}{Hotel Bookings} datasets from Kaggle, along with the corresponding hints (\textit{weak}, \textit{strong}) provided for each.
\subsubsection{Titantic}
\textbf{Errors:}
\begin{enumerate}
    \item The values of the column \texttt{Sex} of $50\%$ of female survivors (identified by Miss. or Mrs. in their names) are randomly changed to male. This constitutes a \textit{categorical shift}.
    \item The values of the column \texttt{Age} are assigned unrealistically low ages (between 2 and 8 years old) for $50\%$ of married female non-survivors. This constitutes a \textit{numerical shift}.
    \item Reduces the values of the column \texttt{Fare} to $10\%$ of their original amounts for passengers with high status titles such as Dr. or Lady in their names. This constitutes a \textit{numerical shift}.
\end{enumerate}

\textbf{Hints (weak, strong):}
\begin{itemize}
    \item "Errors are in the \texttt{Sex}, \texttt{Age} and \texttt{Fare} columns."
    \item "Errors are here: Female survisors had their \texttt{sex} entry corrupted, The same happened for the \texttt{age} of female married non-survivors, and the \texttt{fare} of some passengers with high social status was corrupted."
\end{itemize}

\subsubsection{Meat Consumption}
\textbf{Errors:}
\begin{enumerate}
    \item The values for the column \texttt{Poultry} are set to near-zero (random values between 0 and 0.1) for all \texttt{countries} in specific \texttt{years} (1986, 1990, 1993, 1995, 2000, 2005, 2010, 2015), simulating missing or drastically reduced data for those periods. This constitutes a \textit{numerical shift}.
    \item The values for the columns \texttt{Fish and Seafood} consumption for landlocked \texttt{countries}—Afghanistan, Burkina Faso, Chad, Burundi, Central African Republic, Niger, Nepal, Mali, Tajikistan, Uzbekistan, and Kyrgyzstan—are set between the 85th and 95th percentiles of the dataset for all years, creating unrealistically high fish and seafood consumption that does not match real-world patterns. This constitutes a \textit{numerical shift}.
    \item The values for the total meat consumption (\texttt{Poultry, Beef, Sheep and goat, Pork, Other meats, Fish and seafood}) for Mauritius, Italy, Japan, Vietnam, China, and Mexico are progressively increased by $30\%$ each \texttt{year} between 1997 and 2004, resulting in unrealistically high meat consumption sums during these years. This constitutes a \textit{numerical shift}.
\end{enumerate}

\textbf{Hints (weak, strong):}
\begin{itemize}
    \item "Errors are observed in 1) certain \texttt{years} [1986, 1990, 1993, 1995, 2000, 2005, 2010, 2015]), 2) In some \texttt{countries} regarding \texttt{fish and seafood} consumption, and in consecutive \texttt{years} for the following countries Mauritius, Italy, Japan, Vietnam, China, Mexico."
    \item "Observed errors:
    \begin{enumerate}
        \item In the \texttt{years} [1986, 1990, 1993, 1995, 2000, 2005, 2010, 2015], \texttt{poultry} consumption is significantly underreported.
        \item In landlocked \texttt{countries} such as Afghanistan, Burkina Faso, Chad, Burundi, Central African Republic, Niger, Nepal, Mali, Tajikistan, Uzbekistan, and Kyrgyzstan, \texttt{fish and seafood consumption} is reported to be excessively high.
        \item In \texttt{countries} like Mauritius, Italy, Japan, Vietnam, China, and Mexico, the total meat consumption is notably overreported during the \texttt{years} [1997, 1998, 1999, 2000, 2001, 2003, 2004]."
    \end{enumerate}
\end{itemize}

\subsubsection{Hotel Bookings}
\textbf{Errors:}
\begin{enumerate}
    \item The values for the column \texttt{lead\_time} are increased by 10 days for all records in the \texttt{year} 2016, resulting in consistently inflated booking \texttt{lead times} for that \texttt{year}. This constitutes a \textit{numerical shift}.
    \item Changes the \texttt{deposit\_type} to \texttt{Non Refund} for all bookings made through the TA/TO \texttt{distribution channel} in 2017, creating an unrealistic scenario where all such \texttt{reservations} are non-refundable. This constitutes a \textit{categorical shift}.
    \item Replaces $70\%$ of the \texttt{country} entries with NaN values in the \texttt{country} column to PRT (Portugal) for all years except 2015, although \verb|PRT| is the most frequent country based on the available data from 2015. This constitutes a \textit{NaN corruption}.
\end{enumerate}
\textbf{Hints (weak, strong):}
\begin{itemize}
    \item "Errors are in the \texttt{lead\_time}, \texttt{deposit} and \texttt{country} columns, there are no errors in any entries from 2015."
    \item "Errors are here: There is a systematic bias in the \texttt{lead\_time} of 2016, the deposit with \texttt{distribution\_channel} TA/TO looks wrong in 2017 and often when people arrive from PRT, the \texttt{country} is not recorded."
\end{itemize}

\newpage
\subsection{Initial prompt $P_{0}$ provided to the model}\label{sec:initial_prompt}
\begin{lstlisting}
You have been provided with a dataset containing structural errors that degrade your model's performance. Your goal is to identify and correct these errors so that the model's F1 score increases from {metadata["performance_test"]} to at least {goal}. The target column is {metadata["target_column"]}.

1. Locate and fix any inconsistent or illogical values in the data. Examples of inconsistencies might include negative values where they cannot logically occur, incorrect category assignments, or strange drift in the data.

2. Your training dataset is located in sandbox/train.csv.

3. Each time you attempt a fix, save your cleaned dataset as train_cleaned_v*.csv with an updated version number (e.g., train_cleaned_v1.csv, train_cleaned_v2.csv, etc.).

4. After updating and saving the cleaned dataset, call submit_clean_data() to evaluate the model on the test set.

5. You may run code to explore and fix the dataset with execute_code_ipython_shell(), but you cannot:
- Plot graphs or use graphical libraries.
- Add new columns to the dataset.
- Modify or drop the _competition_index column (doing so will break the evaluation process).
- Add new rows to the dataset.

6. Focus on increasing the model's F1 score by correcting structural issues. These issues typically involve values that are inconsistent with the dataset's real-world context.

Tip: You need to understand the dataset to identify and correct errors. Don't just submit attempts blindly.

The dataset description is as follows: 
{dataset_description}

This is the preprocessing pipeline code before training:
```python
{metadata["pipeline_code"]}
```

Hint: {Hints}
\end{lstlisting}

Where metadata["pipeline\_code"] is dataset specific
\newpage

\end{document}